\documentclass[10pt,twocolumn,letterpaper]{article}

\usepackage{cvpr}
\usepackage{times}
\usepackage{epsfig}
\usepackage{graphicx}
\usepackage{amsmath}
\usepackage{amssymb}

\usepackage{bm}
\usepackage{subfigure}
\usepackage{amsthm}
\usepackage{multirow,array}
\usepackage{color}
\usepackage{algorithmic,algorithm}
\usepackage{lipsum}
\usepackage{booktabs}

\usepackage{color, colortbl}
\definecolor{LightCyan}{rgb}{0.9,0.9,0.9}

\usepackage[breaklinks=true,bookmarks=false]{hyperref}

\cvprfinalcopy 

\def\cvprPaperID{****} 

\ifcvprfinal\fi
\begin{document}

\title{Revisiting Local Descriptor based Image-to-Class Measure \\ for Few-shot Learning}


\author{Wenbin Li\textsuperscript{$1$},\quad Lei Wang\textsuperscript{$2$},\quad Jinglin Xu\textsuperscript{$3$},\quad Jing Huo\textsuperscript{$1$},\quad Yang Gao\textsuperscript{$1$},\quad Jiebo Luo\textsuperscript{$4$} \\
\textsuperscript{$1$}Nanjing University, China,\quad
\textsuperscript{$2$}University of Wollongong, Australia\\
\textsuperscript{$3$}Northwestern Polytechnical University, China,\quad
\textsuperscript{$4$}University of Rochester, USA\\
}

\maketitle

\begin{abstract}
Few-shot learning in image classification aims to learn a classifier to classify images when only few training examples are available for each class. Recent work has achieved promising classification performance, where an image-level feature based measure is usually used. In this paper, we argue that a measure at such a level may not be effective enough in light of the scarcity of examples in few-shot learning. Instead, we think a local descriptor based image-to-class measure should be taken, inspired by its surprising success in the heydays of local invariant features. Specifically, building upon the recent episodic training mechanism, we propose a Deep Nearest Neighbor Neural Network (DN4 in short) and train it in an end-to-end manner. Its key difference from the literature is the replacement of the image-level feature based measure in the final layer by a local descriptor based image-to-class measure. This measure is conducted online via a $k$-nearest neighbor search over the deep local descriptors of convolutional feature maps. The proposed DN4 not only learns the optimal deep local descriptors for the image-to-class measure, but also utilizes the higher efficiency of such a measure in the case of example scarcity, thanks to the exchangeability of visual patterns across the images in the same class. Our work leads to a simple, effective, and computationally efficient framework for few-shot learning. Experimental study on benchmark datasets consistently shows its superiority over the related state-of-the-art, with the largest absolute improvement of $17\%$ over the next best. The source code can be available from \UrlFont{https://github.com/WenbinLee/DN4.git}.
\end{abstract}

\section{Introduction}
Few-shot learning aims to learn a model with good generalization capability such that it can be readily adapted to new unseen classes (concepts) by accessing only one or few examples. However, the extremely limited number of examples per class can hardly represent the class distribution effectively, making this task truly challenging.

To tackle the few-shot learning task, a variety of methods have been proposed, which can be roughly divided into two types, \emph{i.e.,} meta-learning based~\cite{SantoroBBWL16,ravi2017optimization,mishra2018simple} and metric-learning based~\cite{koch2015siamese,SnellSZ17Prototypical,yang2018learning}. The former type introduces a meta-learning paradigm~\cite{thrun1998lifelong,VilaltaD02} to learn an across-task meta-learner for generalizing to new unseen tasks. They usually resort to recurrent neural networks or long short term memory networks to learn a memory network~\cite{WestonCB14,miller2016key} to store knowledge. The latter type adopts a relatively simpler architecture to learn a deep embedding space to transfer representation (knowledge). This type of methods usually relies on the metric learning and episodic training mechanism~\cite{VinyalsBLKW16}. Both types of methods have greatly advanced the development of few-shot learning.

These existing methods mainly focus on making knowledge transfer~\cite{VinyalsBLKW16,cai2018memory}, concept representation~\cite{SnellSZ17Prototypical,garcia2018few} or relation measure~\cite{yang2018learning}, but have not paid sufficient attention to the way of the final classification. They generally take the common practice, \emph{i.e.,} using the image-level pooled features or fully connected layers designed for larger-scale image classification, for the few-shot case. Considering the unique characteristic of few-shot learning (\emph{i.e.,} the scarcity of examples for each training class), such a common practice may not be appropriate any more.

In this paper, we revisit the Naive-Bayes Nearest-Neighbor (NBNN) approach~\cite{boiman2008defense} published a decade ago, and investigate its effectiveness in the context of the latest few-shot learning research. The NBNN approach demonstrated a surprising success when the bag-of-features model with local invariant features (\emph{i.e.,} SIFT) was popular. That work provides two key insights. First, summarizing the local features of an image into a compact image-level representation could lose considerable discriminative information. It will not be recoverable when the number of training examples is small. Second, in this case, directly using these local features for classification will not work if an image-to-image measure is used. Instead, an image-to-class measure should be taken, by exploiting the fact that a new image can be roughly ``composed'' using the pieces of other images in the same class. The above two insights inspire us to review the way of the final classification in the existing methods for few-shot learning and reconsider the NBNN approach for this task with deep learning.

Specifically, we develop a novel \emph{Deep Nearest Neighbor Neural Network} (DN4 in short) for few-shot learning. It follows the recent episodic training mechanism and is fully end-to-end trainable. Its key difference from the related existing methods lies in that it replaces the image-level feature based measure in the final layer with a local descriptor based image-to-class measure. Similar to NBNN~\cite{boiman2008defense}, this measure is computed via a $k$-nearest neighbor search over local descriptors, with the difference that these descriptors are now trained deeply via convolutional neural networks. Once trained, applying the proposed network to new few-shot learning tasks is straightforward, consisting of local descriptor extraction and then a nearest neighbor search. Interestingly, in terms of computation, the scarcity of examples per class now turns out to be an ``advantage'' making NBNN more appealing for few-shot learning. It mitigates the computation of searching for the nearest neighbors from a huge set of local descriptors, which is one factor of the lower popularity of NBNN in large-scale image classification.

Experiments are conducted on multiple benchmark datasets to compare the proposed DN4 with the original NBNN and the related state-of-the-art methods for the task of few-shot learning. The proposed method again demonstrates a surprising success. It improves the $1$-shot and $5$-shot accuracy on \emph{mini}ImageNet from $50.44\%$ to $51.24\%$ and from $66.53\%$ to $71.02\%$, respectively. Particularly, on fine-grained datasets it achieves the largest absolute improvement over the next best method by $17\%$.


\begin{figure*}[!tp]
\centering
\includegraphics[width=0.75\textwidth]{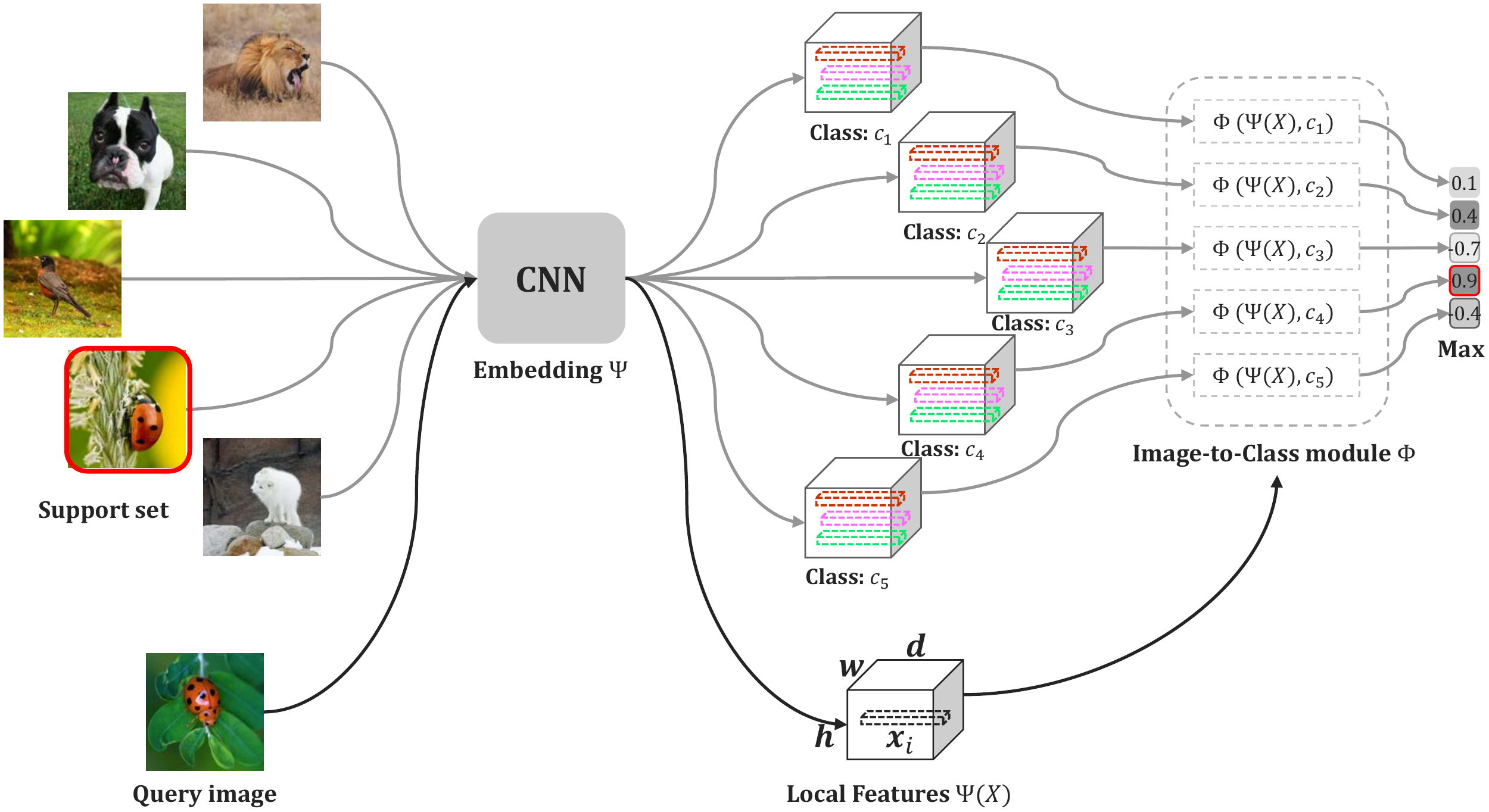}
\caption{Illustration of the proposed Deep Nearest Neighbor Neural Network (DN4 in short) for a few-shot learning task in the $5$-way and $1$-shot setting. As shown, this framework consists of a CNN-based embedding module $\Psi(\cdot)$ for learning deep local descriptors and an image-to-class module $\Phi(\cdot)$ for measuring the similarity between a given query image $X$ and each of the classes, $c_i~(i=1,2,\cdots,5)$.}
\label{fig_flowchart}
\end{figure*}

\section{Related Work}
Among the recent literature of few-shot learning, the transfer learning based methods are most relevant to the proposed method. Therefore, we briefly review two main branches of this kind of methods as follows.

\textbf{Meta-learning based methods.} As shown by the representative work~\cite{SantoroBBWL16,ravi2017optimization,FinnAL17,cai2018memory,gidaris2018dynamic}, the meta-learning based methods train a meta-learner with the meta-learning or the learning-to-learn paradigm~\cite{thrun1998lifelong,thrun1998learning,VilaltaD02} for few-shot learning. This is beneficial for identifying how to update the parameters of the learner's model. For instance, Santoro \emph{et al.}~\cite{SantoroBBWL16} trained an LSTM as a controller to interact with an external memory module. And the work~\cite{ravi2017optimization} adopted an LSTM-based meta-learner as an optimizer to train another classifier as well as learning a task-common initialization for this classifier. The work of MM-Net~\cite{cai2018memory} constructed a contextual learner to predict the parameters of an embedding network for unlabeled images by using memory slots.

Although the meta-learning based methods can achieve excellent results for few-shot classification, it is difficult to train their complicated memory-addressing architecture because of the temporally-linear hidden state dependency~\cite{mishra2018simple}. Compared with the methods in this branch, the proposed framework DN4 can be trained more easily in an end-to-end manner from scratch, \emph{e.g.,} by only using a common single convolutional neural networks (CNN), and could provide quite competitive results.

\textbf{Metric-learning based methods.}
The metric-learning based methods mainly depend on learning an informative similarity metric, as demonstrated by the representative work~\cite{koch2015siamese,VinyalsBLKW16,triantafillou2017few,SnellSZ17Prototypical,garcia2018few,yang2018learning,li2019CovaMNet}. Specifically, to introduce the metric-based method into few-shot learning, Koch~\emph{et al.}~\cite{koch2015siamese} originally utilized a Siamese Neural Network to learn powerful discriminative representations and then generalized them to unseen classes. And then, Vinyals~\emph{et al.}~\cite{VinyalsBLKW16} introduced the episodic training mechanism into few-shot learning and proposed the Matching Nets by combining attention and memory together. In \cite{SnellSZ17Prototypical}, a Prototypical Network was proposed by taking the mean of each class as its corresponding prototype representation to learn a metric space. Recently, Sung~\emph{et al.} considered the relation between query images and class images, and presented a Relation Network~\cite{yang2018learning} to learn a deep non-linear measure.

The proposed framework DN4 belongs to the metric-learning based methods. However, a key difference from them is that the above methods mainly adopt the image-level features for classification, while the proposed DN4 exploits deep local descriptors and the image-to-class measure for classification, as inspired by the NBNN approach~\cite{boiman2008defense}. As will be shown in the experimental part, the proposed DN4 can clearly outperform the several state-of-the-art metric-learning based methods.

\section{The Proposed Method}

\subsection{Problem Formulation}
Let $\mathcal{S}$ denote a support set, which contains $C$ different image classes and $K$ labeled samples per class. Given a query set $\mathcal{Q}$, few-shot learning aims to classify each unlabeled sample in $\mathcal{Q}$ according to the set $\mathcal{S}$. This setting is also called $C$-way $K$-shot classification. Unfortunately, when $\mathcal{S}$ only has few samples per class, it will be hard to effectively learn a model to classify the samples in $\mathcal{Q}$. Usually, the literature resorts to an auxiliary set $\mathcal{A}$ to learn transferable knowledge to improve the classification on $\mathcal{Q}$. Note that the set $\mathcal{A}$ can contain a large number of classes and labeled samples, but it has a \textit{disjoint} class label space with respect to the set $\mathcal{S}$.

The episodic training mechanism~\cite{VinyalsBLKW16} has been demonstrated in the literature as an effective approach to learning the transferable knowledge from $\mathcal{A}$, and it will also be adopted in this work. Specifically, at each iteration, an episode is constructed to train the classification model by simulating a few-shot learning task. The episode consists of a support set $\mathcal{A}_S$ and a query set $\mathcal{A}_Q$ that are randomly sampled from the auxiliary set $\mathcal{A}$. Generally, $\mathcal{A}_S$ has the same numbers of ways (\emph{i.e.,} classes) and shots as $\mathcal{S}$. In other words, there are exactly $C$ classes and $K$ samples per class in $\mathcal{A}_S$. During training, tens of thousands of episodes will be constructed to train the classification model, namely the episodic training. In the test stage, with the support set $\mathcal{S}$, the learned model can be directly used to classify each image in $\mathcal{Q}$.

\subsection{Motivation from the NBNN Approach}
This work is largely inspired by the Naive-Bayes Nearest-Neighbor (NBNN) method in~\cite{boiman2008defense}. The two key observations of NBNN are described as follows, and we show that they apply squarely to few-shot learning.

First, for the (then-popular) bag-of-features model in image classification, local invariant features are usually quantized into visual words to generate the distribution of words (\emph{e.g.}, a histogram obtained by sum-pooling) in an image. It is observed in~\cite{boiman2008defense} that due to quantization error, such an image-level representation could significantly lose discriminative information. If there are sufficient training samples, the subsequent learning process (\emph{e.g.}, via support vector machines) can somehow recover from such a loss, still showing satisfactory classification performance. Nevertheless, when training samples are insufficient, this loss is unrecoverable and leads to poor classification.

Few-shot learning is impacted more significantly by the issue of example scarcity than NBNN. And the existing methods usually pool the last convolutional feature maps (\emph{e.g.}, via the global average pooling or fully connected layer) to an image-level representation for the final classification. In this case, such an information loss will also occur and is unrecoverable.

Second, as further observed in~\cite{boiman2008defense}, using the local invariant features of two images, instead of their image-level representations, to measure an image-to-image similarity for classification will still incur a poor result. This is because such an image-to-image similarity does not generalize beyond training samples. When the number of training samples is small, a query image could be different from any training samples of the same class due to intra-class variation or background clutter. Instead, an image-to-class measure should be used. Specifically, the local invariant features from all training samples in the same class are collected into one pool. This measure evaluates the proximity (\emph{e.g.,} via nearest-neighbor search) of the local features of a query image to the pool of each class for classification.

Again, this observation applies to few-shot learning. Essentially, the above image-to-class measure breaks the boundaries of training images in the same class, and uses their local features collectively to provide a richer and more flexible representation for a class. As indicated in~\cite{boiman2008defense}, this setting can be justified by a fact that a new image can be roughly ``composed'' by using the pieces of other images in the same class (\emph{i.e.}, the exchangeability of visual patterns across the images in the same class).

\subsection{The Proposed DN4 Framework }
The above analysis motivates us to review the way of the final classification in few-shot learning and reconsider the NBNN approach. This leads to the proposed framework \emph{Deep Nearest Neighbor Neural Network} (DN4 in short).

As illustrated in Figure~\ref{fig_flowchart}, DN4 mainly consists of two components: a deep embedding module $\Psi$ and an image-to-class measure module $\Phi$. The former learns deep local descriptors for all images. With the learned descriptors, the latter calculates the aforementioned image-to-class measure. Importantly, these two modules are integrated into a unified network and trained in an end-to-end manner from scratch. Also, note that the designed image-to-class module can readily work with any deep embedding module.

\textbf{Deep embedding module.} The module $\Psi$ routinely learns the feature representations for query and support images. Any proper CNN can be used. Note that $\Psi$ only contains convolutional layers but has no fully connected layer, since we just need deep local descriptors to compute the image-to-class measure. In short, given an image $X$, $\Psi(X)$ will be an $h\!\times\!w\!\times\!d$ tensor, which can be viewed as a set of $m~(m\!=\!hw)$ $d$-dimensional local descriptors as
\begin{equation}\label{fun1}
\Psi(X)=[\bm{x}_1,\ldots,\bm{x}_m]\in\mathbb{R}^{d\times m}\,,
\end{equation}
where $\bm{x}_i$ is the $i$-th deep local descriptor. In our experiments, given an image with a resolution of $84\times 84$, we can get $h=w=21$ and $d=64$. It means that each image has $441$ deep local descriptors in total.

\textbf{Image-to-Class module.} The module $\Phi$ uses the deep local descriptors from all training images in a class to construct a local descriptor space for this class. In this space, we calculate the image-to-class similarity (or distance) between a query image and this class via $k$-NN, as in~\cite{boiman2008defense}.

Specifically, through the module $\Psi$, a given query image $q$ will be embedded as $\Psi(q)=[\bm{x}_1,\ldots,\bm{x}_m]\in\mathbb{R}^{d\times m}$. For each descriptor $\bm{x}_i$, we find its $k$-nearest neighbors $\bm{\hat{x}}_i^j|_{j=1}^k$ in a class $c$.
Then we calculate the similarity between $\bm{x}_i$ and each
$\bm{\hat{x}}_i$, and sum the $mk$ similarities as the image-to-class similarity between $q$ and the class $c$. Mathematically, the image-to-class measure can be easily expressed as
\begin{equation}\label{fun2}
\begin{split}
&\Phi\big(\Psi(q), c\big)=\sum_{i=1}^{m}\sum_{j=1}^k \mathrm{cos}(\bm{x}_i,\bm{\hat{x}}_i^j) \\
&\mathrm{cos}(\bm{x}_i,\bm{\hat{x}}_i)=\frac{\bm{x}_i^\top \bm{\hat{x}}_i}{\|\bm{x}_i\|\cdot\|\bm{\hat{x}}_i\|}\,,
\end{split}
\end{equation}
where $\mathrm{cos}(\cdot)$ indicates the cosine similarity. Other similarity or distance functions can certainly be employed.

Note that in terms of computational efficiency, the image-to-class measure seems more suitable for few-shot classification than the generic image classification focused in~\cite{boiman2008defense}. The major computational issue in NBNN caused by searching for $k$-nearest neighbors from a huge pool of local descriptors has now been substantially weakened due to the much smaller number of training samples in few-shot setting. This makes the proposed framework computationally efficient. Furthermore, compared with NBNN, it will be more promising, by benefiting from the deep feature representations that are much more powerful than the hand-crafted features used in NBNN.

Finally, it is worth mentioning that the image-to-class module in DN4 is non-parametric. So the entire classification model is non-parametric if not considering the embedding module $\Psi$. Since a non-parametric model does not involve parameter learning, the over-fitting issue in parametric few-shot learning methods (\emph{e.g.,} learning a fully connected layer over image-level representation) can also be mitigated to some extent.

\subsection{Network Architecture}
For fair comparison with the state-of-the-art methods, we take a commonly used four-layer convolutional neural network as the embedding module. It contains four convolutional blocks, each of which consists of a convolutional layer, a batch normalization layer and a Leaky ReLU layer. Besides, for the first two convolutional blocks, an additional $2\times2$ max-pooling layer is also appended, respectively. This embedding network is named \emph{Conv-64F}, since there are $64$ filters of size $3\times 3$ in each convolutional layer. As for the image-to-class module, the only hyper-parameter is the parameter $k$, which will be discussed in the experiment.

At each iteration of the episodic training, we feed a support set $\mathcal{S}$ and a query image $q$ into our model. Through the embedding module $\Psi$, we obtain all the deep local representations for all these images. Then via the module $\Phi$, we calculate the image-to-class similarity between $q$ and each class by Eq.~(\ref{fun2}). For a $C$-way $K$-shot task, we can get a similarity vector $\bm{z}\in\mathbb{R}^C$. The class corresponding to the largest component of $\bm{z}$ will be the prediction for $q$.

\section{Experimental Results}
\label{section_4}
The main goal of this section is to investigate two interesting questions: (1) How does the pre-trained deep features based NBNN without episodic training perform on the few-shot learning? (2) How does our proposed DN4 framework, \emph{i.e.,} a CNN based NBNN in an end-to-end episodic training manner, perform on the few-shot learning?

\subsection{Datasets}
We conduct all the experiments on four benchmark datasets as follows.

\textbf{\emph{mini}ImageNet.} As a mini-version of ImageNet~\cite{RussakovskyDSKS15}, this dataset~\cite{VinyalsBLKW16} contains $100$ classes with $600$ images per class, and has a resolution of $84\times84$ for each image. Following the splits used in \cite{ravi2017optimization}, we take $64$, $16$ and $20$ classes for training (auxiliary), validation and test, respectively.

\textbf{Stanford Dogs.} This dataset~\cite{khosla2011novel} is originally used for the task of fine-grained image classification, including $120$ breeds (classes) of dogs with a total number of $20,580$ images. Here, we conduct fine-grained few-shot classification task on this dataset, and take $70$, $20$ and $30$ classes for training (auxiliary), validation and test, respectively.

\textbf{Stanford Cars.} This dataset~\cite{krause20133d} is also a benchmark dataset for fine-grained classification task, which consists of $196$ classes of cars with a total number of $16,185$ images. Similarly, $130$, $17$ and $49$ classes in this dataset are split for training (auxiliary), validation and test.

\textbf{CUB-200.} This dataset~\cite{WelinderEtal2010} contains $6033$ images from $200$ bird species. In a similar way, we select $130$, $20$ and $50$ classes for training (auxiliary), validation and test.

For the last three fine-grained datasets, all the images in these datasets are resized to $84\times84$ as \emph{mini}ImageNet.
\begin{table*}[!tp]\small
	\centering
    \tabcolsep=18pt
	\caption{The mean accuracies of the $5$-way $1$-shot and $5$-shot tasks on the \emph{mini}ImageNet dataset, with $95\%$ confidence intervals. The second column refers to which kind of embedding module is employed, \emph{e.g.,} \emph{Conv-32F} and \emph{Conv-64F} \emph{etc}. The third column denotes the type of this method, \emph{i.e.,} meta-learning based or metric-learning based. $^\ast$ Results reported by the original work. $^\ddag$ Results re-implemented in the same setting for a fair comparison.}
	\begin{tabular}{@{}lcccc@{}}
	\toprule
    \multirow{2}{*}{\textbf{Model}} &\multirow{2}{*}{\textbf{Embedding}} &\multirow{2}{*}{\textbf{Type}}  &\multicolumn{2}{c}{\textbf{5-Way Accuracy (\%)}}　\\
    \cmidrule{4-5}
                                    &           &               &1-shot    &5-shot                                                 \\
    \midrule
	\textbf{$k$-NN (Deep global features)}                         &{\small\emph{Conv-64F}}   &Metric   & $27.23${\scriptsize$\pm1.41$}     & $49.29${\scriptsize$\pm1.56$}   \\
	\textbf{NBNN (Deep local features)}                            &{\small\emph{Conv-64F}}   &Metric   & $44.10${\scriptsize$\pm1.17$}     & $58.84${\scriptsize$\pm1.10$}   \\
    \midrule
	\textbf{Matching Nets FCE}$^\ast$~\cite{VinyalsBLKW16}  &{\small\emph{Conv-64F}}   &Metric   & $43.56${\scriptsize$\pm0.84$}     & $55.31${\scriptsize$\pm0.73$}    \\
    \textbf{Prototypical Nets}$^\ddag$~\cite{SnellSZ17Prototypical} &{\small\emph{Conv-64F}}  &Metric
                                                                                              & $48.45${\scriptsize$\pm0.96$}        & $66.53${\scriptsize$\pm0.51$}  \\
    \textbf{Prototypical Nets}$^\ast$~\cite{SnellSZ17Prototypical} &{\small\emph{Conv-64F}}   &Metric
                                                                                                 & $49.42${\scriptsize$\pm0.78$}     & $68.20${\scriptsize$\pm0.66$}  \\
    \textbf{GNN}$^\ddag$~\cite{garcia2018few}               &{\small\emph{Conv-64F}}   &Metric   & $49.02${\scriptsize$\pm0.98$}     & $63.50${\scriptsize$\pm0.84$}    \\
    \textbf{GNN}$^\ast$~\cite{garcia2018few}                &{\small\emph{Conv-256F}}  &Metric   & $50.33${\scriptsize$\pm0.36$}     & $66.41${\scriptsize$\pm0.63$}    \\
    \textbf{Relation Net}$^\ast$~\cite{yang2018learning}    &{\small\emph{Conv-64F}}   &Metric   & $50.44${\scriptsize$\pm0.82$}     & $65.32${\scriptsize$\pm0.70$}    \\

    \midrule
     \\[-1em]
	\textbf{Our DN4} ($k$=3)             &{\small\emph{Conv-64F}}     &Metric  & $\bm{51.24}${\scriptsize$\bm{\pm0.74}$}  &$\bm{71.02}${\scriptsize$\bm{\pm0.64}$}  　\\
    \\[-1em]
	\midrule
	\\[-1.4em]
	\multicolumn{5}{c}{To take a whole picture of the-state-of-art methods} \\
	\\[-1.4em]
	\midrule
	Meta-Learner LSTM$^\ast$~\cite{ravi2017optimization} &{\small\emph{Conv-32F}}      &Meta  & $43.44${\scriptsize$\pm0.77$}  & $60.60${\scriptsize$\pm0.71$}    \\
    SNAIL$^\ast$~\cite{mishra2018simple}                 &{\small\emph{Conv-32F}}      &Meta  & $45.10$                        & $55.20$     \\
    MAML$^\ast$~\cite{FinnAL17}                          &{\small\emph{Conv-32F}}      &Meta  & $48.70${\scriptsize$\pm1.84$}  & $63.11${\scriptsize$\pm0.92$}    \\
    MM-Net$^\ast$~\cite{cai2018memory}                   &{\small\emph{Conv-64F}}      &Meta  & $53.37${\scriptsize$\pm0.48$}  & $66.97${\scriptsize$\pm0.35$}    \\
    SNAIL$^\ast$~\cite{mishra2018simple}                 &{\small\emph{ResNet-256F}}   &Meta  & $55.71${\scriptsize$\pm0.99$}  & $68.88${\scriptsize$\pm0.92$}     \\
    Dynamic-Net$^\ast$~\cite{gidaris2018dynamic}         &{\small\emph{ResNet-256F}}   &Meta  & $55.45${\scriptsize$\pm0.89$}  & $70.13${\scriptsize$\pm0.68$}    \\
    Dynamic-Net$^\ast$~\cite{gidaris2018dynamic}         &{\small\emph{Conv-64F}}      &Meta  & $56.20${\scriptsize$\pm0.86$}  & $72.81${\scriptsize$\pm0.62$}    \\
    \bottomrule
	\end{tabular}
	\label{miniImageNet_result}
\end{table*}

\begin{table*}[!tbp]\small
	\centering
	 \tabcolsep=5pt
	\caption{The mean accuracies of the $5$-way $1$-shot and $5$-shot tasks on three fine-grained datasets, \emph{i.e., Stanford Dogs, Stanford Cars and CUB-200}, with $95\%$ confidence intervals. For each setting, the best and the second best methods are highlighted.}
	\begin{tabular}{@{}lccccccc@{}}
	\toprule
     \multirow{3}{*}{\textbf{Model}} &\multirow{3}{*}{\textbf{Embed.}}  &\multicolumn{6}{c}{\textbf{5-Way Accuracy (\%)}}　     \\
     \cmidrule{3-8}
            &      &\multicolumn{2}{c}{\emph{Stanford Dogs}}　&\multicolumn{2}{c}{\emph{Stanford Cars}} &\multicolumn{2}{c}{\emph{CUB-200}}\\
            &      &1-shot   &5-shot                         &1-shot     &5-shot                 &1-shot      &5-shot         \\
    \midrule
	\textbf{$k$-NN (Deep global features)}                                            &{\small\emph{Conv-64F}}
	                                &$26.14${\scriptsize$\pm0.91$}             &$43.14${\scriptsize$\pm1.02$}       &$23.50${\scriptsize$\pm0.88$}
	                                &$34.45${\scriptsize$\pm0.98$}             &$25.81${\scriptsize$\pm0.90$}       &$45.34${\scriptsize$\pm1.03$} \\
	\textbf{NBNN (Deep local features)}                                              &{\small\emph{Conv-64F}}
	                                &$31.42${\scriptsize$\pm1.12$}             &$42.17${\scriptsize$\pm0.99$}       &$28.18${\scriptsize$\pm1.24$}
	                                &$38.27${\scriptsize$\pm0.92$}              &$35.29${\scriptsize$\pm1.03$}       &$47.97${\scriptsize$\pm0.96$} \\
    \midrule
	\textbf{Matching Nets FCE}$^\ddag$~\cite{VinyalsBLKW16}                            &{\small\emph{Conv-64F}}
	                                &$35.80${\scriptsize$\pm0.99$}             &$47.50${\scriptsize$\pm1.03$}       &$34.80${\scriptsize$\pm0.98$}
                                    &$44.70${\scriptsize$\pm1.03$}             &$45.30${\scriptsize$\pm1.03$}       &$59.50${\scriptsize$\pm1.01$}\\
    \textbf{Prototypical Nets}$^\ddag$~\cite{SnellSZ17Prototypical}                    &{\small\emph{Conv-64F}}
                                    &$37.59${\scriptsize$\pm1.00$}             &$48.19${\scriptsize$\pm1.03$}       &$40.90${\scriptsize$\pm1.01$}
                                    &$52.93${\scriptsize$\pm1.03$}             &$37.36${\scriptsize$\pm1.00$}       &$45.28${\scriptsize$\pm1.03$} \\
    \textbf{GNN}$^\ddag$~\cite{garcia2018few}                                          &{\small\emph{Conv-64F}}
                                    &$\bm{46.98}${\scriptsize$\bm{\pm0.98}$}   &$62.27${\scriptsize$\pm0.95$}       &$55.85${\scriptsize$\pm0.97$}
                                    &$71.25${\scriptsize$\pm0.89$}             &$\bm{51.83}${\scriptsize$\bm{\pm0.98}$}       &$63.69${\scriptsize$\pm0.94$} \\
    \midrule
	\textbf{Our DN4} ($k$=1)                                              &{\small\emph{Conv-64F}}
	                                &$45.41${\scriptsize$\pm0.76$}             &$\bm{63.51}${\scriptsize$\bm{\pm0.62}$}  &$\bm{59.84}${\scriptsize$\bm{\pm0.80}$}
                                    &$\bm{88.65}${\scriptsize$\bm{\pm0.44}$}   &$46.84${\scriptsize$\pm0.81$}            &$\bm{74.92}${\scriptsize$\bm{\pm0.64}$}\\
    \textbf{Our DN4-DA} ($k$=1)                                           &{\small\emph{Conv-64F}}
                                    &$\bm{45.73}${\scriptsize$\bm{\pm0.76}$}   &$\bm{66.33}${\scriptsize$\bm{\pm0.66}$}   &$\bm{61.51}${\scriptsize$\bm{\pm0.85}$}
                                    &$\bm{89.60}${\scriptsize$\bm{\pm0.44}$}   &$\bm{53.15}${\scriptsize$\bm{\pm0.84}$}   &$\bm{81.90}${\scriptsize$\bm{\pm0.60}$}\\
    \bottomrule
	\end{tabular}
	\label{fineGrained_result}
\end{table*}

\subsection{Experimental Setting}
All experiments are conducted around the $C$-way $K$-shot classification task on the above datasets. To be specific, $5$-way $1$-shot and $5$-shot classification tasks will be conducted on all these datasets. During training, we randomly sample and construct $300,000$ episodes to train all of our models by employing the episodic training mechanism. In each episode, besides the $K$ support images (shots) in each class, $15$ and $10$ query images will also be selected from each class for the $1$-shot and $5$-shot settings, respectively. In other words, for a $5$-way $1$-shot task, there will be $5$ support images and $75$ query images in one training episode. To train our model, we adopt the Adam algorithm~\cite{kingma2015adam} with an initial learning rate of $1\!\times\!10^{-3}$ and reduce it by half of every $100,000$ episodes.

During test, we randomly sample $600$ episodes from the test set, and take the top-$1$ mean accuracy as the evaluation criterion. This process will be repeated five times, and the final mean accuracy will be reported. Moreover, the $95\%$ confidence intervals are also reported. Notably, all of our models are trained from scratch in an end-to-end manner, and do not need fine-tuning in the test stage.

\subsection{Comparison Methods}
\textbf{Baseline methods.} To illustrate the basic classification performance on the above datasets, we implement a baseline method \emph{$k$-NN (Deep global features)}. Particularly, we adopt the basic embedding network \emph{Conv-64F} and append three additional FC layers to train a classification network on the corresponding training (auxiliary) dataset. During test, we use this pre-trained network to extract features from the last FC layer and use a $k$-NN classifier to get the final classification results. Also, to answer the first question at the beginning of Section~\ref{section_4}, we re-implement the NBNN algorithm~\cite{boiman2008defense} by using the pre-trained Conv-64F truncated from the above $k$-NN (Deep global features) method. This new NBNN algorithm employing the deep local descriptors instead of the hand-crafted descriptors (\emph{i.e.,} SIFT), is called \emph{NBNN (Deep local features)}.

\textbf{Metric-learning based methods.} As our method belongs to the metric-learning branch, we mainly compare our model with four state-of-the-art metric-learning based models, including \emph{Matching Nets FCE}~\cite{VinyalsBLKW16}, \emph{Prototypical Nets}~\cite{SnellSZ17Prototypical}, \emph{Relation Net}~\cite{yang2018learning} and \emph{Graph Neural Network} (GNN)~\cite{garcia2018few}. Note that we re-run the GNN model by using the Conv-64F as its embedding module because the original GNN adopts a different embedding module \emph{Conv-256F}, which also has four convolutional layers but with $64$, $96$, $128$ and $256$ filters for the corresponding layers, respectively. Also, we re-run the Prototypical Nets via the same $5$-way training setting instead of the $20$-way training setting in the original work for a fair comparison.

\textbf{Meta-learning based methods.} Besides the metric-learning based models, five state-of-the-art meta-learning based models are also picked for reference. These models include \emph{Meta-Learner LSTM}~\cite{ravi2017optimization}, \emph{Model-agnostic Meta-learning (MAML)}~\cite{FinnAL17}, \emph{Simple Neural AttentIve Learner (SNAIL)}~\cite{mishra2018simple}, \emph{MM-Net}~\cite{cai2018memory} and \emph{Dynamic-Net}~\cite{gidaris2018dynamic}. As SNAIL adopts a much more complicated \emph{ResNet-256F} (a smaller version of ResNet~\cite{he2016deep}) as its embedding module, we will additionally report its results based on the \emph{Conv-32F} provided in its appendix for a fair comparison. Note that Conv-32F has the same architecture with Conv-64F, but with $32$ filters per convolutional layer, which has also been employed by Meta-Learner LSTM and MAML to reduce over-fitting.

\subsection{Few-shot Classification}
The generic few-shot classification task is conducted on \emph{mini}ImageNet. The results are reported in Table~\ref{miniImageNet_result}, where the hyper-parameter $k$ is set as $3$. From Table~\ref{miniImageNet_result}, it is amazing to see that NBNN (Deep local features) can achieve much better results than $k$-NN (Deep global features), and it is even better than Matching Nets FCE, Meta-Learner LSTM and SNAIL (Conv-32F). This not only verifies that the local descriptors can perform better than the image-level features (\emph{i.e.,} FC layer features used by $k$-NN), but also shows that the image-to-class measure is truly promising. However, NBNN (Deep local features) still has a large performance gap compared with the state-of-the-art Prototypical Nets, Relation Net and GNN. The reason is that, as a lazy learning algorithm, NBNN (Deep local features) does not have a training stage and also lacks the episodic training. So far, the first question has been answered.

On the contrary, our proposed DN4 embeds the image-to-class measure into a deep neural network, and can learn the deep local descriptors jointly by employing the episodic training, which indeed obtains superior results. Compared with the metric-learning based models, our DN4 (Conv-64F) gains $7.68\%$, $2.22\%$, $2.79\%$ and $0.8\%$ improvements over Matching Nets FCE, GNN$^\ddag$ (Conv-64F), Prototypical Nets$^\ddag$ (\emph{i.e.,} via $5$-way training setting) and Relation Net on the $5$-way $1$-shot classification task, respectively. On the $5$-way $5$-shot classification task, we can even get $15.71\%$, $7.52\%$, $4.49\%$ and $5.7\%$ significant improvements over these models. The reason is that these methods usually use image-level features whose number is too small, while our DN4 adopts learnable deep local descriptors which are more abundant especially in the $5$-shot setting. On the other hand, local descriptors enjoy the exchangeability characteristic, making the distribution of each class built upon the local descriptors more effective than the one built upon the image-level features. Therefore, the second question can also be answered.

To take a whole picture of the few-shot learning area, we also report the results of the state-of-the-art meta-learning based methods. We can see that our DN4 is still competitive with these methods. Especially in the $5$-way $5$-shot setting, our DN4 gains $15.82\%$, $10.42\%$, $7.91\%$ and $4.05\%$ improvements over SNAIL (Conv-32F), Meta-Learner LSTM, MAML and MM-Net, respectively. As for the Dynamic-Net, a two-stage model, it pre-trains its model with all classes together before conducting the few-shot training, while our DN4 does not. More importantly, our DN4 only has one single unified network, which is much simpler than these meta-learning based methods with additional complicated memory-addressing architectures.

\subsection{Fine-grained Few-shot Classification}
Besides the generic few-shot classification, we also conduct fine-grained few-shot classification tasks on three fine-grained datasets, \emph{i.e.,} Stanford Dogs, Stanford Cars and CUB-200. Two baseline models and three state-of-the-art models are implemented on these three datasets, \emph{i.e.,} $k$-NN (Deep global features), NBNN (Deep local features), Matching Nets FCE~\cite{VinyalsBLKW16}, Prototypical Nets~\cite{SnellSZ17Prototypical} and GNN~\cite{garcia2018few}. The results are shown in Table~\ref{fineGrained_result}. In general, the fine-grained few-shot classification task is more challenging than the generic one due to the smaller inter-class and larger intra-class variations of the fine-grained datasets. It can be seen by comparing the performance of the same methods between Tables~\ref{miniImageNet_result} and \ref{fineGrained_result}. The performance of the $k$-NN (Deep global features), NBNN (Deep local features) and Prototypical Nets on the fine-grained datasets is worse than that on \emph{mini}ImageNet. It can also be observed that NBNN (Deep local features) performs consistently better than $k$-NN (Deep global features).

Due to the small inter-class variation of the fine-grained task, we choose $k\!=\!1$ for our DN4 to avoid introducing noisy visual patterns. From Table~\ref{fineGrained_result}, we can see that our DN4 performs surprisingly well on these datasets under the $5$-shot setting. Especially on the Stanford Cars, our DN4 gains the largest absolute improvement over the second best method, \emph{i.e.,} GNN, by $17\%$. Under the $1$-shot setting, our DN4 does not perform as well as in the $5$-shot setting. The key reason is that our model relies on the $k$-nearest neighbor algorithm, which is a lazy learning algorithm and its performance depends largely on the number of samples. This characteristic has been shown in Table~\ref{shot_num}, \emph{i.e.,} the performance of DN4 gets better and better as the number of shots increases. Another reason is that these fine-grained datasets are not sufficiently large (\emph{e.g.,} CUB-200 only has $6033$ images), resulting in over-fitting when training deep networks.

To avoid over-fitting, we perform data augmentation on the training (auxiliary) sets by cropping and horizontally flipping randomly. Then, we re-train our model, \emph{i.e.,} DN4-DA, on these augmented datasets but test on the original test sets. It can be observed that our DN4-DA can obtain nearly the best results for both $1$-shot and $5$-shot tasks. The fine-grained recognition largely relies on the subtle local visual patterns, and they can be naturally captured by the learnable deep local descriptors emphasized in our model.

\subsection{Discussion}
\textbf{Ablation study.} To further verify that the image-to-class measure is more effective than the image-to-image measure, we perform an ablation study by developing two image-to-image (IoI for short) variants of DN4. Specifically, the first variant named DN4-IoI-1 concatenates all local descriptors of an image as a high-dimensional ($h \times w\times d$) feature vector and uses the image-to-image measure. As for the second variant (DN4-IoI-2 for short), it keeps the local descriptors like DN4 without concatenation. The only difference between DN4-IoI-2 and DN4 is that DN4-IoI-2 restricts the search for the $k$-NN of a query's local descriptor within each individual support image, while DN4 can search from one entire support class. Under the $1$-shot setting, DN4-IoI-2 is identical with DN4. Both variants still adopt the $k$-NN search, and use $k=1$ and $k=3$ for $1$-shot setting and $5$-shot setting, respectively.

The results on \emph{mini}ImageNet are reported in Table~\ref{ablation_study}. As seen, DN4-IoI-1 performs clearly the worst by using the concatenated global features with the image-to-image measure. In contrast, DN4-IoI-2 performs excellently on both $1$-shot and $5$-shot tasks, which verifies the importance of local descriptors and the exchangeability (within one image). Notably, DN4 is superior to DN4-IoI-2 on the $5$-shot task, which shows that utilizing the exchangeability of visual patterns within a class indeed helps to gain performance.

\begin{table}[!tp]\small
	\centering
    \tabcolsep=14pt
	\caption{The results of the ablation study on \emph{mini}ImageNet.}
	\begin{tabular}{@{}lcc@{}}
	\toprule
    \multirow{2}{*}{\textbf{Model}}   &\multicolumn{2}{c}{\textbf{5-Way Accuracy (\%)}}　\\
    \cmidrule{2-3}
                                      &1-shot           &5-shot               \\
    \midrule
    \textbf{DN4-IoI-1}               & $37.39${\scriptsize$\pm0.82$}            &$50.47${\scriptsize$\pm0.66$}  　\\
    \textbf{DN4-IoI-2}               & $51.14${\scriptsize$\pm0.79$}  &$69.52${\scriptsize$\pm0.62$}  　\\
	\textbf{DN4}                     & $\bm{51.24}${\scriptsize$\bm{\pm0.74}$}  &$\bm{71.02}${\scriptsize$\bm{\pm0.64}$}  　\\
    \bottomrule
	\end{tabular}
	\label{ablation_study}
\end{table}
\begin{table}[!tp]\small
	\centering
    \tabcolsep=9pt
	\caption{The $5$-way $5$-shot mean accuracy (\%) of our DN4 by varying the value of $k\in\{1,3,5,7\}$ during training on \emph{mini}ImageNet.}
	\begin{tabular}{@{}lcccc@{}}
	\toprule
    \multirow{2}{*}{\textbf{Model}}  &\multicolumn{4}{c}{\textbf{5-way 5-shot Accuracy (\%)}}　\\
    \cmidrule{2-5}
                      & $k=1$          & $k=3$          & $k=5$        & $k=7$      \\
    \midrule
    \textbf{DN4}      & $\bm{71.95}$   & $71.02$        & $70.20$      & $68.56$    \\
    \bottomrule
	\end{tabular}
	\label{k_num}
\end{table}


\textbf{Influence of backbone networks.} Besides the commonly used \emph{Conv-64F}, we also evaluate our model by using another deeper embedding module, \emph{i.e.,} \emph{ResNet-256F} used by SNAIL~\cite{mishra2018simple} and Dynamic-Net~\cite{gidaris2018dynamic}. The details of ResNet-256F can refer to SNAIL~\cite{mishra2018simple}. When using ResNet-256F as the embedding module, the accuracy of DN4 reaches $54.37\pm0.36\%$ for the $5$-way $1$-shot task and $74.44\pm0.29\%$ for the $5$-shot task. As seen, with a deeper backbone network, DN4 can perform better than the case of using the shallow Conv-64F. Moreover, when using the same ResNet-256F as the embedding module, our DN4 (ResNet-256F) can gain $4.31\%$ improvements over Dynamic-Net (ResNet-256F) (\emph{i.e.,} $70.13\pm0.68\%$) under the $5$-shot setting (see Table~\ref{miniImageNet_result}).

\textbf{Influence of neighbors.}
In the image-to-class module, we need to find the $k$-nearest neighbors in one support class for each local descriptor of a query image. Next, we measure the image-to-class similarity between a query image and a specific class. How to choose a suitable hyper-parameter $k$ is thus a key. For this purpose, we perform a $5$-way $5$-shot task on \emph{mini}ImageNet by varying the value of $k\in\{1,3,5,7\}$, and show the results in Table~\ref{k_num}. It can be seen that the value of $k$ has a mild impact on the performance. Therefore, in our model, $k$ should be selected according to the specific task.
\begin{table}[!tp]\small
	\centering
    \tabcolsep=9pt
	\caption{The $5$-way $K$-shot mean accuracy (\%) of our DN4 by varying the number of shots ($K\!=\!{1,2,3,4,5}$) during training on \emph{mini}ImageNet. For each test setting, the best result is highlighted.}
	\begin{tabular}{@{}lccccc@{}}
	\toprule
    \multirow{2}{*}{\textbf{Train}}  &\multicolumn{5}{c}{\textbf{Test}}　\\
    \cmidrule{2-6}
                    & 1-shot                     & 2-shot               & 3-shot                & 4-shot               & 5-shot    \\
    \midrule
    1-shot          & $\underline{51.24}$        & $58.13$              & $62.10$               & $64.22$              & $66.10$    \\
    2-shot          & $50.69$                    & $\underline{58.53}$  & $62.31$               & $64.84$              & $66.49$    \\
    3-shot          & $53.22$                    & $60.74$              & $\underline{64.95}$   & $67.52$              & $69.35$    \\
    4-shot          & $52.43$                    & $60.90$              & $65.33$               & $\underline{67.93}$  & $69.70$    \\
    5-shot          & $\bm{53.85}$  & $\bm{61.78}$         & $\bm{66.16}$          & $\bm{68.92}$         & $\underline{\bm{71.02}}$    \\
    \bottomrule
	\end{tabular}
	\label{shot_num}
\end{table}

\textbf{Influence of shots.}
The episodic training mechanism is popular in current few-shot learning methods. The basic rule is the matching condition between training and test. It means that, in the training stage, the numbers of ways and shots should keep consistent with those adopted in the test stage. In other words, if we want to perform a $5$-way $1$-shot task, the same $5$-way $1$-shot setting should be maintained in the training stage. However, in the real training stage, we still want to know the influence of mismatching conditions, \emph{i.e.,} under-matching condition and over-matching condition. We find that the over-matching condition can achieve better performance than the matching condition, and much better than the under-matching condition.

Basically, for the under-matching condition, we use a smaller number of shots in the training stage, and conversely, use a larger number of shots for the over-matching condition. We fix the number of ways but vary the number of shots during training to learn several different models. Then we test these models under different shot settings, where the number of shots is changed but the number of ways is fixed. A $5$-way $K$-shot ($K\!=\!1,2,3,4,5$) task is conducted on \emph{mini}ImageNet by using our DN4. The results are presented in Table~\ref{shot_num}, where the entries on the diagonal are the results of the matching condition. The results in the upper triangle are the results of the under-matching condition. Also, the lower triangle contains the results of the over-matching condition. It can be seen that the results in the lower triangle are better than those on the diagonal, and the results on the diagonal are better than those in the upper triangle. This exactly verifies our statement made above. It is also worth mentioning that if we use a $5$-shot trained model and test it on the $1$-shot task, we can obtain an accuracy of $53.85\%$. This result is quite high in this task, and much better than $51.24\%$ obtained by the $1$-shot trained model using our DN4 under a matching condition.

\begin{figure}[!tp]
\centering
\subfigure[NBNN]{
          \includegraphics[width=0.15\textwidth]{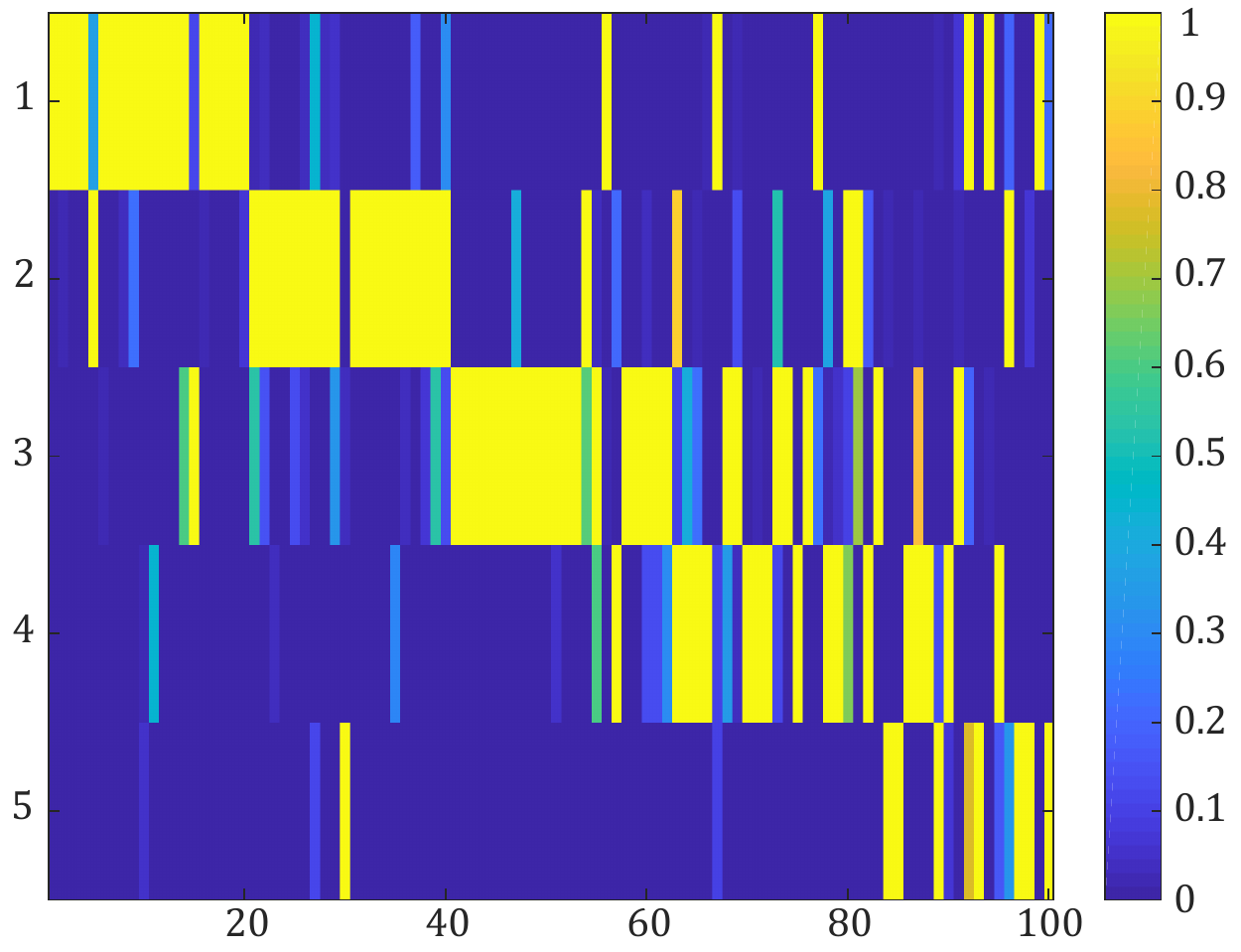}}
\subfigure[Our DN4]{
          \includegraphics[width=0.15\textwidth]{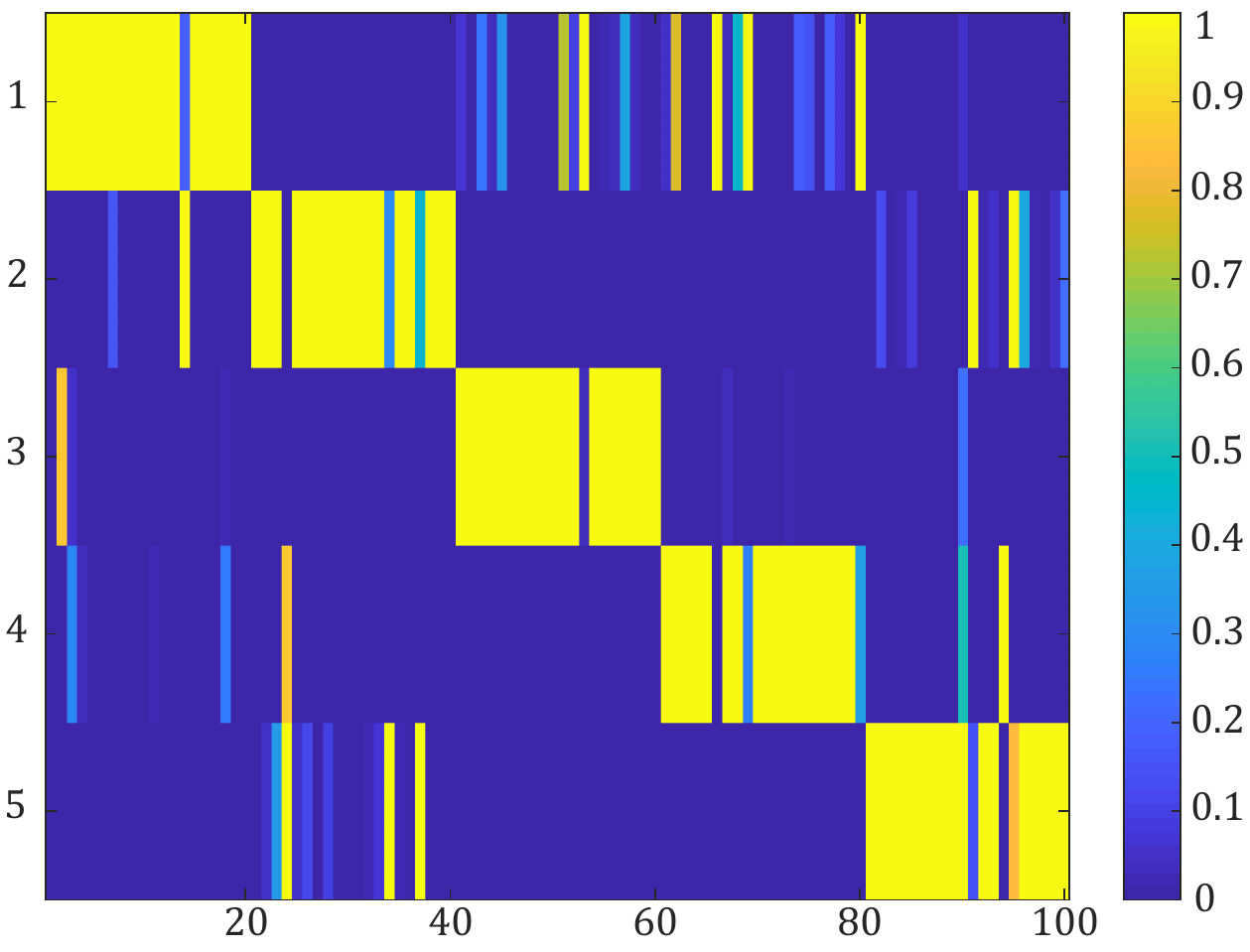}}
\subfigure[Ground Truth]{
          \includegraphics[width=0.15\textwidth]{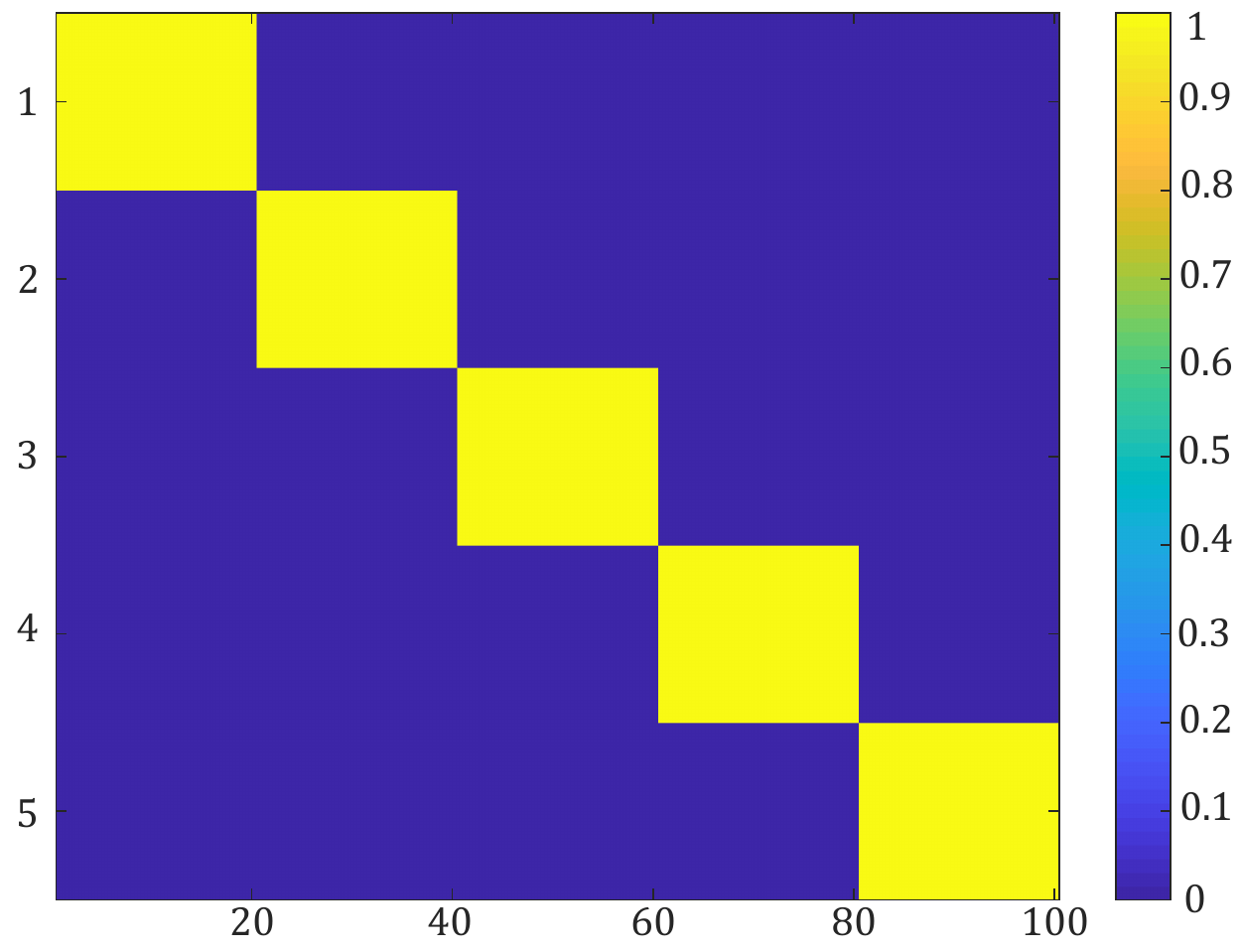}}
\caption{Similarity matrices of NBNN (Deep local Features), our DN4 and the ground truth on \emph{mini}ImageNet under the $5$-way $5$-shot setting. Vertical axis denotes the five classes in the support set. Horizontal axis denotes $20$ query images per class. The warmer colors indicate higher similarities.}
\label{visualization}
\end{figure}

\textbf{Visualization.}
We visualize the similarity matrices learned by NBNN (Deep local features) and our DN4 under the $5$-way $5$-shot setting on \emph{mini}ImageNet. Both of them are image-to-class measure based models. We select $20$ query images from each class (\emph{i.e.,} $100$ query images in total), calculate the similarity between each query image and each class, and visualize the $5\!\times\!100$ similarity matrices. From Figure~\ref{visualization}, it can be seen that the results of DN4 are much closer to the ground truth than those of NBNN, which demonstrates that the end-to-end manner is more effective.

\textbf{Runtime.}
Although NBNN performs successfully in the literature~\cite{boiman2008defense}, it did not become popular. One key reason is the high computational complexity of the nearest-neighbor search, especially in large-scale image classification tasks. Fortunately, under the few-shot setting our framework can enjoy the excellent performance of NBNN without being significantly affected by its computational issue. Generally, during training for a $5$-way $1$-shot or $5$-shot task, one episode (batch) time is $0.31$s or $0.38$s with $75$ or $50$ query images on a single Nvidia GTX $1080$Ti GPU and a single Intel i$7$-$3820$ CPU. During test, it will be more efficient, and only takes $0.18$s for one episode. Moreover, the efficiency of our model can be further improved with optimized parallel implementation.

\section{Conclusions}
In this paper, we revisit the local descriptor based image-to-class measure and propose a simple and effective Deep Nearest Neighbor Neural Network (DN4) for few-shot learning. We emphasize and verify the importance and value of the learnable deep local descriptors, which are more suitable than image-level features for the few-shot problem and can well boost the classification performance. We also verify that the image-to-class measure is superior to the image-to-image measure, owing to the exchangeability of visual patterns within a class.

\section*{Acknowledgements}
This work is partially supported by the NSF awards (Nos. 1704309, 1722847, 1813709), National NSF of China (Nos. 61432008, 61806092), Jiangsu Natural Science Foundation (No. BK20180326), the Collaborative Innovation Center of Novel Software Technology and Industrialization, and Innovation Foundation for Doctor Dissertation of Northwestern Polytechnical University (No. CX201814).

{\small
\bibliographystyle{ieee}
\bibliography{references}

\begin{thebibliography}{10}\itemsep=-1pt

\bibitem{boiman2008defense}
O.~Boiman, E.~Shechtman, and M.~Irani.
\newblock In defense of nearest-neighbor based image classification.
\newblock In {\em CVPR}, pages 1--8. IEEE, 2008.

\bibitem{cai2018memory}
Q.~Cai, Y.~Pan, T.~Yao, C.~Yan, and T.~Mei.
\newblock Memory matching networks for one-shot image recognition.
\newblock In {\em CVPR}, pages 4080--4088, 2018.

\bibitem{FinnAL17}
C.~Finn, P.~Abbeel, and S.~Levine.
\newblock Model-agnostic meta-learning for fast adaptation of deep networks.
\newblock In {\em ICML}, pages 1126--1135, 2017.

\bibitem{garcia2018few}
V.~Garcia and J.~Bruna.
\newblock Few-shot learning with graph neural networks.
\newblock {\em ICLR}, 2018.

\bibitem{gidaris2018dynamic}
S.~Gidaris and N.~Komodakis.
\newblock Dynamic few-shot visual learning without forgetting.
\newblock In {\em CVPR}, pages 4367--4375, 2018.

\bibitem{he2016deep}
K.~He, X.~Zhang, S.~Ren, and J.~Sun.
\newblock Deep residual learning for image recognition.
\newblock In {\em CVPR}, pages 770--778, 2016.

\bibitem{khosla2011novel}
A.~Khosla, N.~Jayadevaprakash, B.~Yao, and F.-F. Li.
\newblock Novel dataset for fine-grained image categorization: Stanford dogs.
\newblock In {\em CVPR Workshop}, volume~2, page~1, 2011.

\bibitem{kingma2015adam}
D.~P. Kingma and J.~Ba.
\newblock Adam: A method for stochastic optimization.
\newblock {\em ICLR}, 2015.

\bibitem{koch2015siamese}
G.~Koch, R.~Zemel, and R.~Salakhutdinov.
\newblock Siamese neural networks for one-shot image recognition.
\newblock In {\em ICML Workshop}, volume~2, 2015.

\bibitem{krause20133d}
J.~Krause, M.~Stark, J.~Deng, and L.~Fei-Fei.
\newblock 3d object representations for fine-grained categorization.
\newblock In {\em ICCV Workshop}, pages 554--561, 2013.

\bibitem{li2019CovaMNet}
W.~Li, J.~Xu, J.~Huo, L.~Wang, G.~Yang, and J.~Luo.
\newblock Distribution consistency based covariance metric networks for
  few-shot learning.
\newblock In {\em AAAI}, 2019.

\bibitem{miller2016key}
A.~Miller, A.~Fisch, J.~Dodge, A.-H. Karimi, A.~Bordes, and J.~Weston.
\newblock Key-value memory networks for directly reading documents.
\newblock {\em EMNLP}, 2016.

\bibitem{mishra2018simple}
N.~Mishra, M.~Rohaninejad, X.~Chen, and P.~Abbeel.
\newblock A simple neural attentive meta-learner.
\newblock {\em ICLR}, 2018.

\bibitem{ravi2017optimization}
S.~Ravi and H.~Larochelle.
\newblock Optimization as a model for few-shot learning.
\newblock {\em ICLR}, 2017.

\bibitem{RussakovskyDSKS15}
O.~Russakovsky, J.~Deng, H.~Su, J.~Krause, S.~Satheesh, S.~Ma, Z.~Huang,
  A.~Karpathy, A.~Khosla, M.~S. Bernstein, A.~C. Berg, and F.~Li.
\newblock Imagenet large scale visual recognition challenge.
\newblock {\em IJCV}, 115(3):211--252, 2015.

\bibitem{SantoroBBWL16}
A.~Santoro, S.~Bartunov, M.~Botvinick, D.~Wierstra, and T.~P. Lillicrap.
\newblock Meta-learning with memory-augmented neural networks.
\newblock In {\em ICML}, pages 1842--1850, 2016.

\bibitem{SnellSZ17Prototypical}
J.~Snell, K.~Swersky, and R.~S. Zemel.
\newblock Prototypical networks for few-shot learning.
\newblock In {\em NIPS}, pages 4080--4090, 2017.

\bibitem{thrun1998lifelong}
S.~Thrun.
\newblock Lifelong learning algorithms.
\newblock In {\em Learning to learn}, pages 181--209. Springer, 1998.

\bibitem{thrun1998learning}
S.~Thrun and L.~Pratt.
\newblock Learning to learn: Introduction and overview.
\newblock In {\em Learning to learn}, pages 3--17. Springer, 1998.

\bibitem{triantafillou2017few}
E.~Triantafillou, R.~Zemel, and R.~Urtasun.
\newblock Few-shot learning through an information retrieval lens.
\newblock In {\em NIPS}, pages 2255--2265, 2017.

\bibitem{VilaltaD02}
R.~Vilalta and Y.~Drissi.
\newblock A perspective view and survey of meta-learning.
\newblock {\em AIR}, 18(2):77--95, 2002.

\bibitem{VinyalsBLKW16}
O.~Vinyals, C.~Blundell, T.~Lillicrap, K.~Kavukcuoglu, and D.~Wierstra.
\newblock Matching networks for one shot learning.
\newblock In {\em NIPS}, pages 3630--3638, 2016.

\bibitem{WelinderEtal2010}
P.~Welinder, S.~Branson, T.~Mita, C.~Wah, F.~Schroff, S.~Belongie, and
  P.~Perona.
\newblock {Caltech-UCSD Birds 200}.
\newblock Technical Report CNS-TR-2010-001, California Institute of Technology,
  2010.

\bibitem{WestonCB14}
J.~Weston, S.~Chopra, and A.~Bordes.
\newblock Memory networks.
\newblock {\em ICLR}, 1410.3916, 2015.

\bibitem{yang2018learning}
F.~S.~Y. Yang, L.~Zhang, T.~Xiang, P.~H. Torr, and T.~M. Hospedales.
\newblock Learning to compare: Relation network for few-shot learning.
\newblock {\em CVPR}, 2018.

\end{thebibliography}
}

\end{document}